\PassOptionsToPackage{table,xcdraw}{xcolor}
\documentclass[acmsmall,screen,nonacm]{acmart}

\usepackage{amsmath}
\usepackage{algorithm}
\usepackage{subcaption}
\usepackage[table]{xcolor}
\usepackage[noend]{algpseudocode}
\usepackage{multirow}

\AtBeginDocument{%
  \providecommand\BibTeX{{%
    \normalfont B\kern-0.5em{\scshape i\kern-0.25em b}\kern-0.8em\TeX}}}

\setcopyright{acmcopyright}
\copyrightyear{2021}
\acmYear{2021}
\acmDOI{XXXXXXXXXXXXXXXX}

\acmJournal{TIOT}
\acmVolume{37}
\acmNumber{4}
\acmArticle{111}
\acmMonth{8}

\begin{document}

\title{Exploiting Multi-modal Contextual Sensing for City-bus's Stay Location Characterization: Towards Sub-60 Seconds Accurate Arrival Time Prediction}

\author{Ratna Mandal}
\authornote{All three authors contributed equally to this research.}
\email{ratna.mandal.iem@gmail.com}
\affiliation{%
  \institution{Institute of Engineering \& Management, Kolkata}
  \country{India}
}
\author{Prasenjit Karmakar}
\authornotemark[1]
\email{prasenjitkarmakar52282@gmail.com}
\affiliation{%
  \institution{ACM Member}
  \country{India}
}

\author{Soumyajit Chatterjee}
\authornotemark[1]
\affiliation{%
  \institution{Indian Institute of Technology Kharagpur}
  \country{India}
}
\email{soumyachat@iitkgp.ac.in}

\author{Debaleen Das Spandan}
\affiliation{%
  \institution{BCET Durgapur}
  \country{India}
}
\email{ddsmegh4@gmail.com}

\author{Shouvit Pradhan}
\affiliation{%
 \institution{BCET Durgapur}
 \country{India}
 }
 \email{shaw8wit@gmail.com}

\author{Sujoy Saha}
\affiliation{%
  \institution{National Institute of Technology Durgapur}
  \country{India}}
\email{sujoy.saha@cse.nitdgp.ac.in}

\author{Sandip Chakraborty}
\affiliation{%
  \institution{Indian Institute of Technology Kharagpur}
  \country{India}}
\email{sandipc@cse.iitkgp.ac.in}

\author{Subrata Nandi}
\affiliation{%
  \institution{National Institute of Technology Durgapur}
  \country{India}}
\email{subrata.nandi@cse.nitdgp.ac.in}

\newcommand\todosoumya[1]{\textcolor{red}{#1}}
\newcommand\notesc[1]{\textcolor{blue}{#1}}
\renewcommand{\shortauthors}{Mandal, et al.}

\begin{abstract}
%Intelligent city transportation systems are a core infrastructure of any smart-city. The true ingenuity of such infrastructure lies in providing the commuters with real-time information, allowing them to pre-plan the travel. However, providing preliminary information for transportation systems like public buses in real-time is inherently challenging because of the diverse nature of stay-locations where public buses stop. Although factors extracted from unimodal sources look erratic, a thorough analysis of public bus trajectory for $720$ km of travel at Durgapur, India, reveals that contextual features can characterize these locations more accurately. Accordingly, we develop~\emph{BuStop}, a system for characterizing the stay locations from multi-modal sensing using smartphones. With multi-modal sensing, ~\emph{BuStop} extracts a set of granular features that allow the system to differentiate among the different stay-location types. A thorough analysis of ~\emph{BuStop} using the in-house dataset indicates that the system can accurately identify different types of stay locations. Additionally, we develop a PoC system on top of ~\emph{BuStop} to analyze the framework's potential in predicting arrival time at any given bus-stop. Subsequent analysis of the PoC system, through simulation over the test dataset, shows that characterizing stay-locations helps make accurate arrival time predictions with deviations less than $60$ seconds.
Intelligent city transportation systems are one of the core infrastructures of a smart city. The true ingenuity of such an infrastructure lies in providing the commuters with real-time information about citywide transports like public buses, allowing her (him) to pre-plan the travel. However, providing prior information for transportation systems like public buses in real-time is inherently challenging because of the diverse nature of different stay-locations that a public bus stops. Although straightforward factors stay duration, extracted from unimodal sources like GPS, at these locations look erratic, a thorough analysis of public bus GPS trails for $720$km of bus-travels at the city of Durgapur, a semi-urban city in India, reveals that several other fine-grained contextual features can characterize these locations accurately. Accordingly, we develop~\emph{BuStop}, a system for extracting and characterizing the stay locations from multi-modal sensing using commuters' smartphones. Using this multi-modal information~\emph {BuStop} extracts a set of granular contextual features that allow the system to differentiate among the different stay-location types. A thorough analysis of~\emph{BuStop} using the collected in-house dataset indicates that the system works with high accuracy in identifying different stay locations like regular bus-stops, random ad-hoc stops, stops due to traffic congestion stops at traffic signals, and stops at sharp turns. Additionally, we also develop a proof-of-concept setup on top of~\emph{BuStop} to analyze the potential of the framework in predicting expected arrival time, a critical piece of information required to pre-plan travel, at any given bus-stop. Subsequent analysis of the PoC framework, through simulation over the test dataset, shows that characterizing the stay-locations indeed helps make more accurate arrival time predictions with deviations less than $60$ seconds from the ground-truth arrival time.
\end{abstract}




\maketitle

\section{Introduction}
\label{intro}
Intelligent public transportation has been one of the primary goals behind the development of smart-city infrastructure. Various smartphone-based apps are widely used throughout multiple cities worldwide to assist commuters with the public city-transportation system. An October 2016 survey report by the California Department of Transportation has indicated that people preferred to use public transportation rather than driving while using multi-modal transit planning apps on their smartphones, albeit the United States is predominantly a car-dependent society~\cite{shaheen2016mobile}. Such an advantage comes from utilizing real-time information captured through on-vehicle sensors and Internet-of-Thing (IoT) devices deployed on the vehicles, as well as sharing the commuters' information through smartphone-based apps. However, many-a-times, such systems' accuracy gets hampered, primarily because of various on-road dynamics, like traffic congestion near a market and its domino effect on different interconnecting roads. Although real-time information can help the commuters know the bus's current position on which she has planned to board, it certainly affects her travel planning and time management. As shown in~\cite{firstcite}, commuters do not prefer to spend time outside, waiting for an incoming bus. Therefore, there is a need for an accurate tracking system for city transportation systems.

Traditionally, tracking and localization of vehicles have highly relied on GPS and cellular networks~\cite{trogh2019outdoor}. However, although such unimodal sensing can provide information about the location and state of mobility, i.e., whether moving or stopped, they do not reveal the true nature of the stay-location or, in general, why the vehicle has stopped. This, in turn, has a huge impact on the final information of the expected arrival time required to pre-plan a trip. This problem further intensifies for city transportation systems like public buses, which stop at many points than private or shared cabs. Interestingly, some of the existing works~\cite{fei2019spatiotemporal,meegahapola2019buscope} have tried to explore this problem of characterizing different stay-locations for clearer insight on mobility patterns of city transportation systems like public-buses. Fei \textit{et. al.}~\cite{fei2019spatiotemporal} have analyzed speed patterns around a bus stop location to have a clear insight of traffic congestion or road illumination as random events along the fixed bus-route from odometer data. A similar work~\cite{meegahapola2019buscope} focused on developing a system called \emph{BuSCOPE} to infer the regularity in the trip patterns for individual commuters. \emph{BuSCOPE} can be utilized to predict the Spatio-temporal bus demands in real-time from mobility data generated by smart-card-based trip information of public bus commuters in Singapore city. Although these works' objective is closely related to our problem statement, however, a major limitation of these works is in the unimodal nature of sensing that they rely on. Both of these techniques use sensors like odometer or smart-cards, which make the system less generic. Additionally, these systems can characterize only a subset of stay-locations because of the inherent limitations of these modalities.
 
Accordingly, in this paper, we try to develop a framework that can characterize and detect different stay-locations that can ultimately be used with GPS or cellular tracking to improve the overall tracking systems. We start this by analyzing the different stay-location types that may arise considering the stoppage patterns of public buses in India. From the initial observation, that in general, a public-bus stops at $5$ different types of stay-locations, namely a regular bus-stop, on a traffic signal, due to traffic congestion, while turning on a road, and at random ad-hoc stops to pick additional passengers. Although the inclusion of these diverse sets of stay-locations makes the system much more generic than the existing state-of-the-art applications, characterizing such a diverse set of stay-location types is highly challenging. This is because specific information like stay-duration, which can be easily obtained from unimodal sensing techniques based on GPS, fails to characterize the stay-locations precisely. For example, random ad-hoc stops and stops due to sudden traffic congestion may have highly varying stay-duration patterns. Also, multiple stay-locations can get confounded, which may further aggravate the challenge. For example, a regular bus-stop can be at a major traffic signal; then, depending on the situation or flow of traffic, the characteristics of this stay-location may behave differently.

Interestingly, amidst all these challenges, there lie specific opportunities to characterize these individual stay-location types. One such opportunity is in capturing the overall context surrounding the stay-locations. For example, during traffic congestion, we usually observe much ambient noise generated due to vehicles' cacophony. Similarly, a random ad-hoc stop is generally given in a crowded area where additional passengers may board the bus. Measuring population density over an in-the-wild setup might not be possible; however, factors like WiFi density and the presence of special points-of-interests (PoI) that include landmarks like shopping malls, public religious places, etc., can be good indicators of the same. Notably, some of the previous works like~\cite{38nericell, RoadSoundSense,hornok} have used contextual information through multi-modal sensing analyzing road-conditions and traffic behavior. Therefore, understanding these opportunities, we develop a multi-modal sensing-based framework, named~\emph{BuStop} that tries to capture the overall context surrounding a stay-location and use that information to characterize the same accurately. 

\emph{BuStop} is a machine learning (ML)-based framework to automatically detect different stay-location types for intra-city public bus travels through multi-modal sensing to capture the general contextual information surrounding the stay-location. This paper's primary contribution is exploiting the information from multiple modalities like GPS, microphone, WiFi, inertial sensors (IMU) embedded in common-off-the-shelf (COTS) smartphones, and coupling various fine-grained contextual information like road surface quality, ambient noise, and population density extracted from these sensors. Additionally, the framework also includes static map-based features encoded from publicly-available map services (like Google maps), which allows the framework to take into account the distribution of PoI(s) surrounding a stay-location. This introduction of rich contextual information allows us to discern among various closely related or often confounded stay-locations. Furthermore, in addition to a rich set of contextual features, we also employ judicious feature selection to characterize the stay-locations precisely. We then subsequently use them for training individual models (in an one-vs-all setting) corresponding to each stay-location type. Notably, this approach of using independent models for each stay-location type allows us to detect confounded stay-locations, where the bus has stopped because of more than one reason. From a thorough analysis of the system on the in-house data from $6$ volunteers with a $720$km of bus travel, we observe that the framework can detect the five classes of stay-locations with an average test F\textsubscript{1}-score of $0.83$ ($\pm 0.08$). Besides, we also benchmark this system's performance with a small proof-of-concept (PoC) application that provides the estimated arrival time of a public-bus apriori so that the commuter can plan the travel accordingly.  In summary, our significant contributions are as follows.
\begin{enumerate}
    \item We start by collecting an~\emph{in-house} dataset over the sub-urban city of Durgapur, whereby we collect samples in an open environment during intra-city bus travels by $6$ volunteers for $720$km of bus-travel. Apart from the regular GPS trails, the dataset also contains multi-modal information from sensors like GPS, WiFi, microphone, IMU available in COTS smartphones.
    \item Next, through an extensive study, we identify and derive a set of features that can significantly impact the aggregate bus mobility on a road segment. We identify various features based on trajectory, demography, traffic characteristics, and temporal patterns. Based on these features, we develop classification models to characterize the stay-locations detected from the GPS trails.
    \item Additionally, we benchmark the developed framework by developing a PoC application on top of the developed framework for predicting the arrival times depending on the stoppage patterns. Our analysis shows that the proposed characterization of stay locations helps us having less than one minute error on average in predicting the bus arrival time.  
\end{enumerate}

The remaining paper is arranged as follows. We first start with an extensive literature survey of the recent efforts in digitizing public bus services (see Section~\ref{survey}). Next, we summarise the entire data collection process (Section~\ref{dataset}). We next perform an initial pilot study to understand the primary challenges and subsequent opportunities to help us develop the framework (Section~\ref{pilot}). With the understanding gained from the pilot study, we develop the framework~\emph{BuStop} (Section~\ref{method}). Finally, we evaluate our framework on the in-house collected dataset (Section~\ref{eval}) and discuss the future steps (Section~\ref{discussion}) before concluding the paper.
\section{Related Work}
\label{survey}
There have been a plethora of researches on mobility analysis based on GPS trajectories, many of which focus on intra-city vehicle data analysis~\cite{infocom,RecSys,poonawala2016singapore,raghothama2016analytics,de2015transittrace,stockx2014subwayps,aoki2017early,sinn2012predicting,fei2018spatiotemporal,lee2012http,mazimpaka2016visual,torre2012matching,borresen2017interactive}. Among these, a good number of works have focused on identification and characterization of various point of interests over a route, which can impact the mobility of the vehicles, such as point of intersections~\cite{borresen2017interactive}, various road attributes~\cite{infocom,yin2020multi,verma2017smart,fei2019spatiotemporal}, mobility patterns~\cite{meegahapola2019buscope}, traffic congestion~\cite{kwee2018traffic,hakim2017quantification,raghothama2016analytics}, significant locations on a route~\cite{mazimpaka2016visual}, and so on. A handful of these works also consider intra-city and public bus travels~\cite{meegahapola2019buscope,kwee2018traffic,mazimpaka2016visual,fei2018spatiotemporal,fei2019spatiotemporal,raymond2011location}, where the GPS data has been collected either through a vehicle-mounted sensor or from smartphone crowdsensing. 
In this section, we have presented a comprehensive study of related works in two directions. Section 2.1 presents the motivational survey related to PoI analysis of buses, and section 2.2 explains the literature study regarding the use of additional features and the GPS logs for the study of travel behavior and road surface condition. 

\subsection{Analysis of Bus Stay-locations from Trajectory data}
In~\cite{meegahapola2019buscope}, the authors have developed a system called \emph{BuSCOPE} for real-time predictive analysis of mobility data from the smart-card generated trip information of public bus commuters in Singapore city. They have shown that commuters' aggregated flow and the regularity in the trip patterns for individual commuters can be inferred from such data, which can be utilized to predict the Spatio-temporal bus demands in real-time. However, the only study of passenger's trip data at designated stops limits the framework in characterizing different stay-locations that appear en route. Kwee \textit{et. al.}~\cite{kwee2018traffic} have shown that the public bus trajectory information is good enough to infer the on-route traffic-cascading behaviors that result in a road-congestion. However, this work utilizes only the bus speed derived from trajectory data, which varies upon an individual driver's driving style.

In~\cite{mazimpaka2016visual}, analyzes mobility patterns from bus trajectory data with explicit consideration of different granularity levels of time dimension and locations. However, this study only analyses the GPS data and finds the correlation of road and traffic-related context, which has a significant impact on a bus's mobility pattern. Yan Lyu~\cite{CB-Planner} plans customized bus stops based on the potential travel demands from other real-time travel data sources, such as taxi GPS trajectories and public transport transaction records. They focus on bus passengers' data and analyze only those PoIs where only passengers will be interested in getting on and off the bus. This limits the framework's overall capability in identifying other kinds of stay locations where passengers may not get into the bus but still appears along a bus route. Xiangjie Kong~\cite{SharedBus} predicts travel requirements from time and location-dependent travel behavior of passengers of taxi and shared bus to generate dynamic routes for shared buses.

To understand the charging patterns of electric vehicles for real-time charging schedule, Guang Wang~\cite{bCharge} utilizes the bus stop and mobility data to understand the traffic congestion and different passenger demands, which significantly affects the charging efficiency of electric bus. They are interested in a specific type of PoI, i.e., charging station. In their works, Fei \textit{et al.}~\cite{fei2018spatiotemporal,fei2019spatiotemporal} have used odometer data to infer the impact of various stay locations, like bus-stops, signals, congestion, etc. on the bus routes. Although this work is very similar to our work, there are two differences. First, they have used odometer data, which is less readily available than GPS data; therefore, our proposed framework uses GPS data to infer the stay locations. Second, their work is based on the data from the Washington Metropolitan Area, and therefore, their model based on Dynamic Time Warping fails to capture the uncertainties and randomness during travel, which primarily comes from ad-hoc bus-stops and erratic stop-patterns. Further, their model does not classify different stay locations; instead, they pre-assume the type of stay-locations and analyze their stay durations accordingly.

\begin{table*}
\scriptsize
\caption{Summary of Existing Works}
\label{Survey}
\centering
\begin{tabular}{|m{2.5cm}|m{1.5cm}|m{4.0cm}|m{4.0cm}|}
\hline
\centering \textbf{Frameworks} & \centering \textbf{Sensing}  & \centering \textbf{Objective} & \textbf{Remarks}\\
\hline \hline 
\textbf{BuScope~\cite{meegahapola2019buscope}} & \centering Unimodal & Predicts crowd flow and individual bus-commuting pattern & Only considers smart card data of passengers getting on and off only at bus-stops. No other kinds of stops can be inferred.\\
\hline
\textbf{Traffic-Cascade~\cite{kwee2018traffic}} & \centering  Unimodal & Detects the lifecycle of traffic congestion by extracting congested segments based on speed information of bus trajectory data, and then clustering the extracted segments & Depends only on speed derived from bus trajectory data (i.e. GPS data) which vary with individual driving style.\\
\hline
\textbf{CB-Planner~\cite{CB-Planner}} & \centering Unimodal & Plans a customized bus stops based on the potential travel demands & Makes a bus passenger specific study and analyses only those PoIs of passenger's choice. It does not consider other kinds of stay locations where passengers are not interested to get on and off.\\
\hline
\textbf{Mazimpaka \textit{et.al}~\cite{mazimpaka2016visual}} & \centering Unimodal & Infers the significant locations on a bus-route, like the bus-stops and the stopping patterns, based on trajectory analysis. They have shown that the bus-stops’ stopping patterns depend on space, time, and various other attributes & Restricted to a subset of stay-locations only.\\
\hline
\textbf{Kong \textit{et.al}~\cite{SharedBus}} & \centering Unimodal & Predicts the travel requirement of passengers and plans dynamic route for shuttle bus & Analyses time and location dependant travel behaviour of passengers.\\ 
\hline
\textbf{bCharge~\cite{bCharge}} & \centering Unimodal & Infers the electric bus network to understand its movement and charging patterns & Focus only on a specific type of PoI i.e., charging station for electric buses.\\
\hline
\textbf{Fei \textit{et.al.}~\cite{fei2019spatiotemporal}} & \centering  Unimodal & Infers the impact of various stay locations, like bus-stops, signals, congestion, etc. on the bus routes & Odometer data is not readily available and the work is carried out on the data from the Washington Metropolitan Area, based on Dynamic Time Warping model which fails to capture the uncertainties and randomness during travel.\\
\hline
\textbf{\emph{BuStop}} & \centering Multi-modal & Provide real-time tracking for public-buses for travel pre-planning. & Detects all possible types of stay-locations for a public bus. Uses COTS smartphones as the experimental apparatus.\\
\hline
\end{tabular}
\end{table*}

\subsection{Use of non-GPS features Data for Traffic and Road Surface Analysis}
In addition to GPS logs, researchers have worked on multimodal non-GPS features like video/image, the sound of honks, GSM radio signal, weather data, and road network data, as well as social media feeds for subjective analysis of traffic behavior. Hoang~\cite{Hoang}  utilizes GSM radio signal, road network data, and weather data in addition to GPS logs to capture seasonal changes and instantaneous
changes. Park et al. ~\cite{Park} have used the GPS data in combination with camera images to propose a traffic risk detection model which automatically detects dangerous driving situations by monitoring the driving behavior. Vij et al.
~\cite{vij2018smartphone} have effectively used microphone data of smartphones to identify different traffic states. Authors show that by using audio analysis, detecting a particular traffic state becomes faster and easier. Sen et al. ~\cite{hornok} estimate the speed of a vehicle from vehicular honks. Moreover, in ~\cite{RoadSoundSense} the author presents an acoustic sensing-based technique for real-time congestion monitoring on chaotic roads. Zheng et al. ~\cite{gasconsumption} estimate the gas consumption and pollution levels emitted by vehicles traveling on ways to determine the travel speed along with road segments. Additionally, using road features, PoI, and the road's global position, they aim to infer the traffic volume from travel speed. In another approach, Wang et al. ~\cite{22} combine social media data with road features to identify traffic congestion on city roads. They estimate citywide traffic congestion from social media by mining the road segments' spatial and temporal correlations facing congestion from historical data. The authors extracted the physical features of roads from road network data. They also measured the impact of a social event on nearby road segments. Hence, the above-studied features can be used along with GPS to infer traffic behavior to a certain extent. However, the road surface condition also plays a vital role in affecting the speed of vehicles.

Extensive literature study shows that researchers either have combined online data with sensor data or have combined historical logs of traffic data with smartphone-like GPS, accelerometer data,  microphone data, and mobile phone signals to provide proper road and traffic information. Eriksson ~\cite{Pothole} developed a Pothole  Patrol system to detect road anomalies using accelerometers and  GPS  sensors installed in taxis.  Mohan ~\cite{38nericell} presented a system to monitor road conditions by detecting potholes,  bumps,  honking from an accelerometer, microphone,  GSM  radio signal, and GPS  data.  Vittorio ~\cite{Vittorio} applied the anomaly detection method on mobile devices based on vertical acceleration and  GPS signals.  Perttunen ~\cite{Perttunen} proposed a road anomaly detection method using GPS and acceleration signals.  Wolverine ~\cite{39wolverine} used smartphone sensors to detect road conditions and bumps. The authors have proposed a reorientation using a magnetometer beside the accelerometer and  GPS  sensors. Nuno Silva ~\cite{Silva} detects road condition from  GPS  and accelerometer data. Haofu  Han ~\cite{han} estimates speed from an accelerometer, which senses natural driving conditions in urban environments, including making turns,  stopping, and passing through uneven road surfaces.  They contribute to eliminating the errors in speed estimation caused by accelerations in real-time.  A. Chowdhury ~\cite{Chowdhury} investigated the noise performance of accelerometers available in smartphones and finally apply the analysis for estimating the speed of moving vehicles because sudden changes in vehicle speed are not always captured well by GPS.
\section{Dataset}
\label{dataset}
Understanding the limitations of the existing state-of-the-art systems for digitizing public bus infrastructures, the first observation we gain is that such a system needs to have multi-modal sensing to effectively use various modalities to extract a diverse set of contextual information surrounding a stay-location. However, before diving into the actual framework development, we first look into how and which kind of dataset can be effectively captured considering smartphones as the primary experimental apparatus. For this, we launch an extensive data collection drive summarised as follows.
\begin{figure}
    \centering
    \begin{minipage}{0.50\columnwidth}
        \centering
        \includegraphics[width=0.80\columnwidth,keepaspectratio]{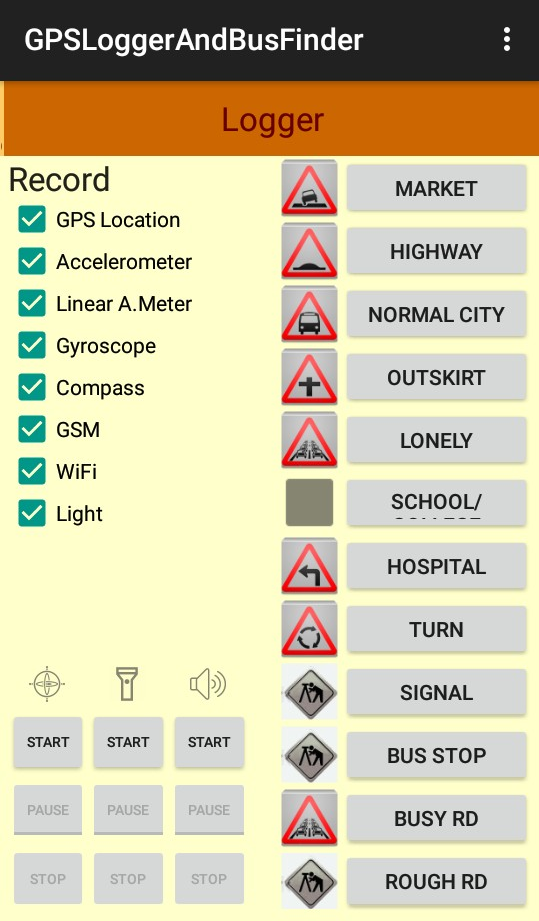}
        \caption{Experimental Apparatus}
        \label{fig:app}
    \end{minipage}\hfill
    \begin{minipage}{0.50\columnwidth}
        \begin{minipage}{\columnwidth}
            \centering
            \includegraphics[width=\columnwidth,keepaspectratio]{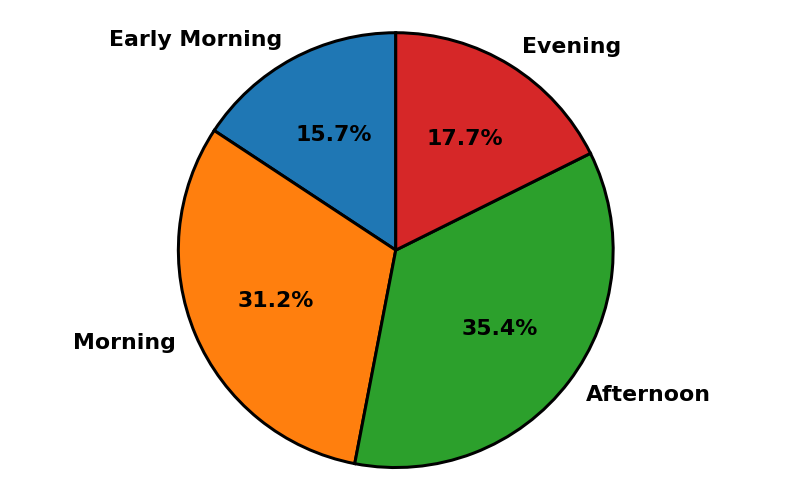}
            \caption{Temporal Distribution of the Dataset}
            \label{fig:temporal_distri}
        \end{minipage}
        
        \begin{minipage}{\columnwidth}
            \centering
            \includegraphics[width=\columnwidth,keepaspectratio]{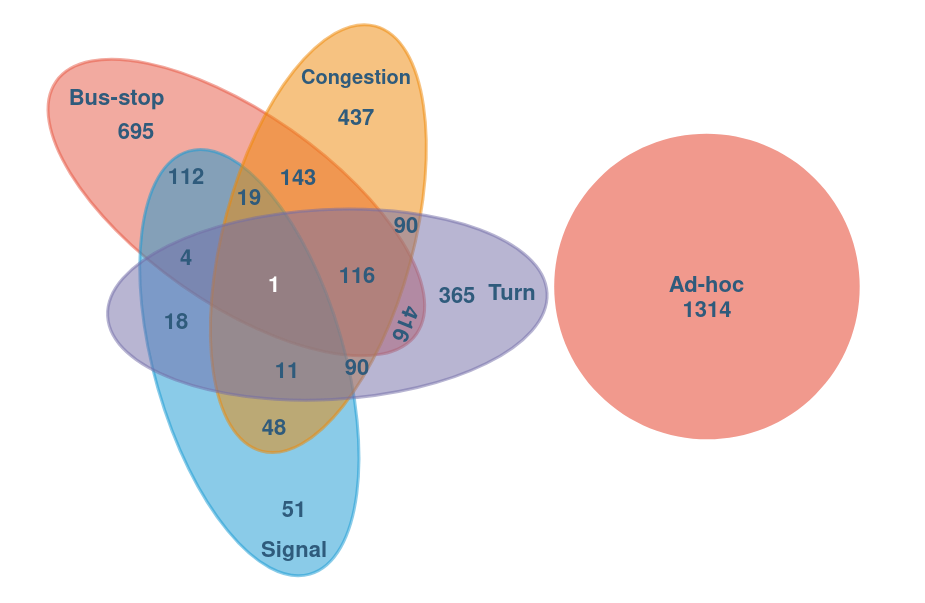}
            \caption{Confounding of Ground Truth}
            \label{fig:dataset_distri_set}
        \end{minipage}
    \end{minipage}
\end{figure}
\subsection{Data Acquisition Framework}
For the data collection, we recruited $6$ volunteers who were asked to travel across the sub-urban city of Durgapur, India, on intra-city buses, while capturing sensor logs through an Android application installed on COTS smartphones ($5$ different builds from a price range of USD $83$\$--$210$\$). The designed Android application acted as the primary experimental apparatus during this entire process of data collection. As shown in Fig.~\ref{fig:app}, the application captured logs from various sensors like GPS (sampled at $1$ Hz), IMU (accelerometer and gyroscope sampled at $197$ Hz), WiFi access point information and microphones (sampled at $8$ kHz) embedded in the smartphones.

Once the initial set of data was collected, we then analyzed the dataset for identifying one or more bus routes with substantial diversity of populous areas, market places, railway stations, etc. Subsequently, we chose a particular route for the remaining data collection, which has the desired diversity for an overall principled analysis of the designed framework and collected the bus trip data over $20$ days using the aforementioned app. In this data collection, each round trip covered a total of $24$km, and the total distance covered during this entire period is $720$km.

Besides the spatial diversities like populous zones, market places, etc., we also captured data across different timezones starting from $6$AM till $9$PM, each day. For this we planned the data collection in different time intervals like -- $6$AM to $9$AM -- \textit{Early Morning}, $9$AM to $1$PM -- \textit{Morning}, $1$PM to $5$PM -- \textit{Afternoon}, and $5$PM to $9$PM -- \textit{Evening}. Fig.~\ref{fig:temporal_distri} shows the overall distribution of data across timezones.

\subsection{Ground-truth Collection}
In addition to the primary data collection, the volunteers also collected ground-truth information that included a manual recording, through the application interface, of the instances when the bus stopped at a stay-location, including the type and reason behind that halt. Besides, we also obtain some of the authoritative information regarding the positions of the bus-stops and signals from the city transport authorities for precise identification of the locations.

Interestingly, during this process, one important observation that we obtained from the ground-truth information concerns the confounding of several stay-location types. As shown in the Venn diagram (see Fig.~\ref{fig:dataset_distri_set}), the stay-location types often get confounded; for example, a bus-stop may occur at the same point as a traffic signal. Understanding this, we specifically asked the volunteers to include all the reasons that may occur because of which the bus has stopped. For all the remaining cases where the volunteers fail to understand why the bus stopped, we exclusively mark them as ad-hoc in the collected dataset.
\section{Pilot Study}
\label{pilot}
Before moving to the actual development of the framework, we first perform a pilot study on the collected dataset to understand the challenges and the opportunities associated with the development. The details of the observations made from the pilot study are as follows. 
\begin{figure}
    \centering
    \begin{minipage}{0.33\columnwidth}
        \centering
        \includegraphics[width=\columnwidth,keepaspectratio]{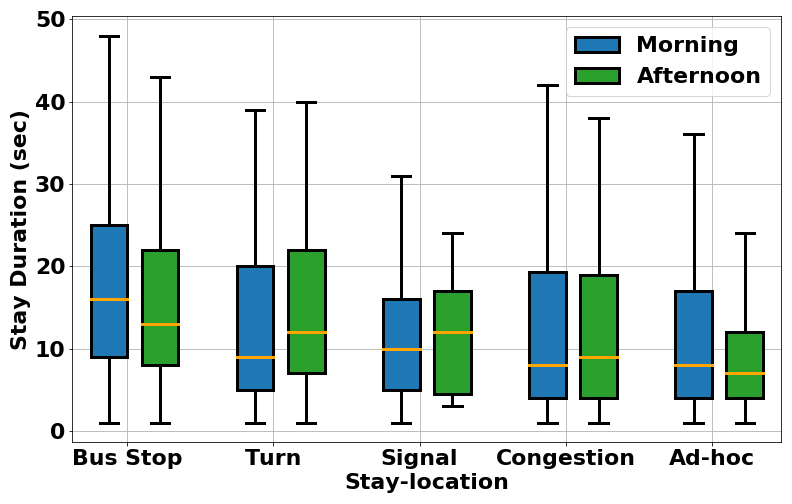}
        \caption{Variations in Stay-duration}
        \label{fig:stay_dur_distri}
    \end{minipage}\hfill
    \begin{minipage}{0.33\columnwidth}
        \centering
        \includegraphics[width=\columnwidth,keepaspectratio]{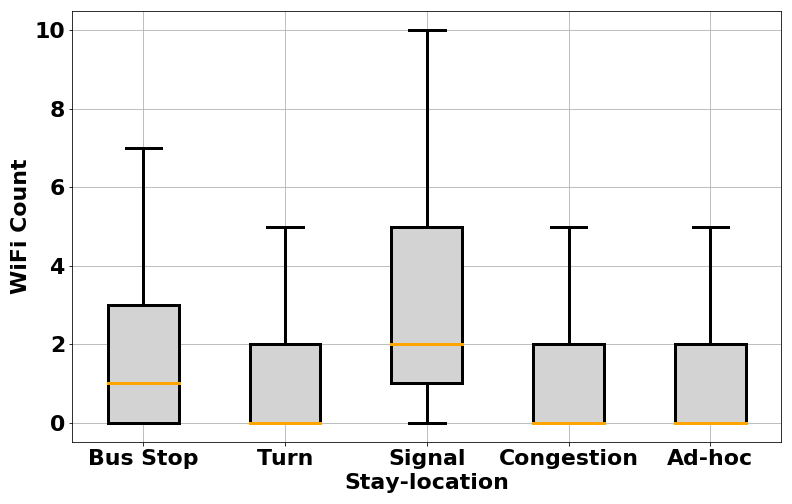}
        \caption{Variations in WiFi Density}
    \label{fig:wifi_distri}
    \end{minipage}\hfill
    \begin{minipage}{0.33\columnwidth}
        \centering
        \includegraphics[width=\columnwidth,keepaspectratio]{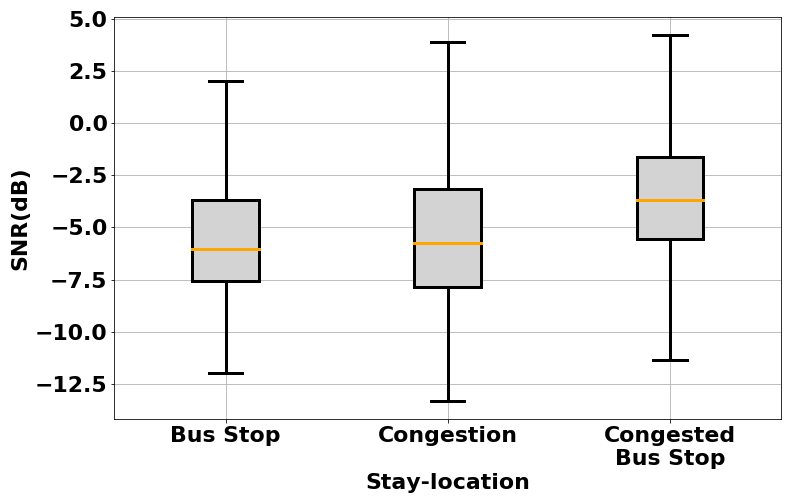}
        \caption{SNR across Stay-locations}
    \label{fig:audio_noise_level}
    \end{minipage}
\end{figure}
\subsection{Challenges}
We first start by analyzing the challenges surrounding the identification of stay-locations to help create a digitized tracker for public bus services. In this context, the first factor that comes into play is the amount of time a bus-stops at any given type of stay-location, also known as stay-duration. From the initial analysis (see Fig.~\ref{fig:stay_dur_distri}) of the distribution of stay-duration for different buses across various stay-location types, we observe that except a very few cases, the stay-duration of a bus does not change significantly across different stay-locations. Further investigation also reveals that although stay-durations are typically fixed for stay-location types like regular bus-stops and signals, however, for ad-hoc locations and the stops made because of congestion, the stay-duration may vary highly. Furthermore, as the different stay-locations may co-occur at the same location, for example a regular bus-stop near a traffic signal, the distribution of stay-duration may vary hugely for a single type of stay-location as well. 

Other than this, another critical factor, as highlighted by previous works like~\cite{own_paper}, can be the inter stay-location distance, which is formally defined as the distance between the consecutive stay-locations. Although this can be a critical feature in characterizing the stay-location, computing this requires future information. This is because new stay-locations may appear due to random factors like congestion or random stops for picking more passengers. Thus, factors like this, albeit essential, may not be practical for real-time monitoring and prediction of stay-locations.
\subsection{Opportunities}
Although specific features like stay-duration may not characterize the stay-locations effectively, however, a closer investigation of these ad-hoc stops reveals that in most cases, such stops are made near places like markets or commercial areas with a high probability of passengers that may vary with days or time intervals of the day. In general, places like markets and commercial areas are also characterized by traffic congestion, which can also contribute to the stops made by a bus. Interestingly, traffic congestion and busy road conditions are usually characterized by the cacophony of horns and vehicles. A typical analysis of traffic congestion, as shown in Fig.~\ref{fig:audio_noise_level}, depicts that the median noise level surrounding traffic congestion is higher than that of a regular bus-stop. Furthermore, we also observe that the SNR is significantly higher for a typically congested bus-stop than in congestion and bus-stop. This is because for a congested bus-stop, in addition to the usual noise of vehicles, there is a steady noise created by the commuters who are planning to board the bus from the location.

Besides, noise levels another significant characteristic of these locations is the overall population density. Although computing such a feature for characterizing a stay-location may not be straightforward and may involve a detailed administrative survey of the area; however, a more straightforward indicator for population density in urban and semi-urban cities can be the overall density of WiFi access points in that area~\cite{buchakchiev2016people}. Motivated by this, we perform a simple pilot study to compute the WiFi density surrounding different stay-locations. As shown in Fig.~\ref{fig:wifi_distri}, demarcated locations like regular bus-stops and traffic signals show high WiFi access-point densities in comparison to random stay-locations appearing due to congestion.
\subsection{Lessons Learned}
From the aforementioned challenges and the subsequent opportunities, we can summarise the lessons learned as follows. We observe that although simple factors like stay-duration, which can be easily extracted from the GPS-based localization and tracking mechanisms, may not be sufficient enough to characterize all the different types of stay-locations. One of the primary reasons behind this is the overall confounding of different stay-locations, thus impacting each other characteristics. 

Interestingly, other than stay-duration and more straightforward factors extracted from basic GPS tracking, several other factors can characterize different stay-locations accurately. For example, factors like ambient noise can be a significant signature to identify traffic congestion precisely. Similarly, population density markers like WiFi density in an area can help characterize the random ad-hoc stops where the buses usually stop to pick up extra passengers to earn additional profit. From the understanding gained through analyzing the challenges and opportunities, we develop the overall framework of~\emph{BuStop} by judicially selecting a set of auxiliary features and then developing a learning model on top of it, as discussed in the next section.
\section{Methodology}
\label{method}
\begin{figure}
    \centering
    \includegraphics[width=0.70\columnwidth,keepaspectratio]{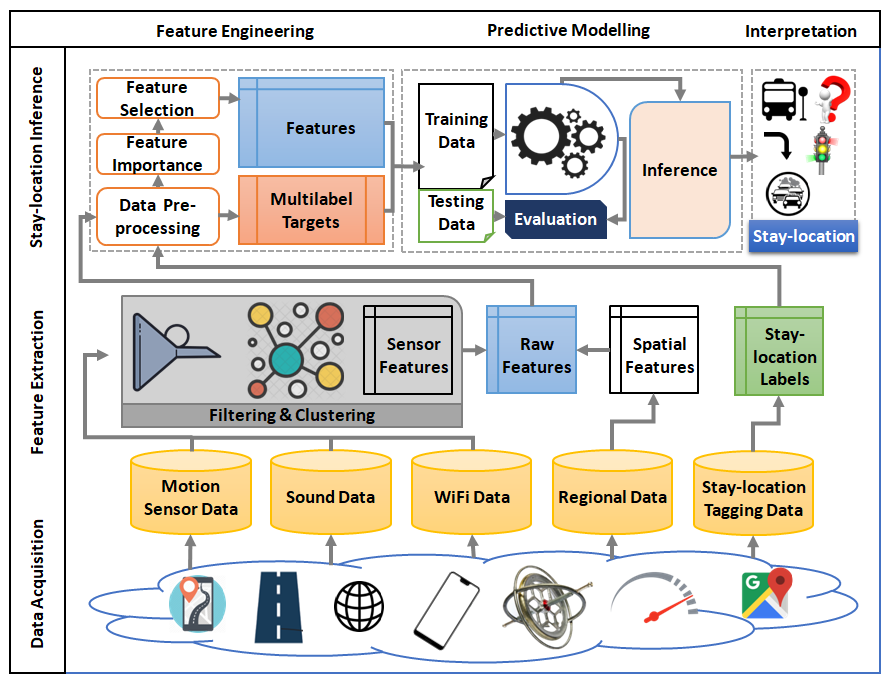}
    \caption{The \emph{BuStop} Framework}
    \label{fig:framework}
\end{figure}

With the lessons learned from the initial pilot study, we develop the framework \emph{BuStop}. As shown in Fig.~\ref{fig:framework}, the overall design has $3$ main components. The first component is the data acquisition framework, which includes the Android application design for data collection (discussed in Section~\ref{dataset}). The second component is the feature extraction module which extracts features that can precisely characterize the context surrounding the different stay-location types. Finally, the last module is the stay-location inference module which exploits these features to train the prediction model. The details follow. 
\subsection{Clustering Zero-speed Points}
Once the raw GPS trails and the data from other sensor modalities are pre-processed, we then extract the~\emph{zero-speed points} from the GPS trail. Formally, a~\emph{zero-speed point} is the point where a vehicle (a public bus in our case) has stopped. In this paper, we extract the zero-speed through a two step process described as follows. In the first step, we filter out the GPS locations where the speed of the vehicle is less than a threshold $\chi$. However, this simple filtering technique generates a cluster of zero-speed points where we observe contiguous values of the redundant GPS locations. This is because the individual zero-speed points may not correspond to independent stay-locations and may be generated because the bus typically stops and moves within a short distance of a zero-speed points due to factors like congestion, traffic, etc. Therefore, we finally, cluster the points within a certain radius $\rho$ into a single-point~\footnote{In this paper $\chi$=$3$m/s and $\rho$=$30$m.} marked as a stay-location.
\subsection{Feature Extraction}
\subsubsection{Stay Duration ($\mathbf{f_1}$)}
The \textit{Stay Duration} ($S_d$) of a public bus at any stay-location is a prominent feature to characterize different categories like traffic signals, and the road turns, traffic congestion, regular bus-stops, and ad-hoc stops. Generally, different stay-locations exhibit different signatures of $S_d$ depending on the time of the day. The count of GPS records, recorded at each stay-location, indicates the total $S_d$ of a bus in seconds at that particular stay-location since the GPS co-ordinates are recorded per-second interval. Besides, we also find this feature to have a reasonable correlation with \textit{Inter Stay-location (or Inter-stop) Distance} between two consecutive stay-locations. In general, we observe that, with high inter-stop distance, $S_d$ increases, and vice versa. Furthermore, we also find that, in different regions of the city, like in densely populated areas like a market, $S_d$ increases due to factors like congestion, increased passenger count, etc. Further, $S_d$ varies significantly depending on the hour of the day. For example, in case of a regular bus-stop, the wait-time is more during the afternoon hours than at night due to the high expectation of getting a passenger.
\subsubsection{Ambient Noise Profile ($\mathbf{f_2}$--$\mathbf{f_6}$)} One of the significant characteristics of road traffic is the amount of noise level in that area. In most cases, busy and congested roads or junctions are characterized by horns' cacophony and noise from vehicles. Understanding this, we consider including the ambient noise profile surrounding the stay-location as a discriminating feature. Although, a straightforward approach of capturing the ambient noise profile can be through the signal-to-noise ratio (SNR) of the acoustic signature surrounding the stay-location (as shown in Section~\ref{pilot}), however, previous works like~\cite{groupsense} have shown that acoustic context can be device dependent and can also be colluded by different factors like distance from the source, presence of muffling objects, etc. Therefore, instead of relying on simple SNR we extract the entire acoustic spectrum for richer information. We do this by computing the~\emph{Mel-frequency Cepstral Coefficients} of the audio sample captured while the bus stopped at the stay-location and then including the top-$5$ coefficients from the sorted list of coefficient values.
\subsubsection{Unique WiFi Hotspot Availability ($\mathbf{f_7}$--$\mathbf{f_8}$)}
Many public spaces like railway stations, offices, malls, and shopping complexes have placed WiFi hotspots for seamless connectivity in the recent past. Additionally, wireless access points are also found in private residential areas as well. With recent endeavors in making smartphones act as WiFi hotspots, the abundance of these access points has increased manifolds. Interestingly, the density of WiFi access points in an area has been observed to be a good indicator of the overall population density~\cite{buchakchiev2016people,jones2007wi}, which can, in turn, be a motivating factor for predicting the stay-location type of a public bus. Therefore, we extract the list of unique WiFi access-points in a stay-point and use the access points' count as a feature. Additionally, we also observe the list of unique WiFi count while the bus moves from one stay-location to another. This additional information also gives us an idea of the commuters' overall flow between the various stay-locations and finally helps us better characterize it for prediction.
\subsubsection{Road Surface Profile ($\mathbf{f_9}$)}
The condition of the road surface is a significant factor for determining the stay location of a bus. Buses typically slow down when the road condition is poor. Poor road conditions in densely populous and busy areas also leads to traffic congestion. As a result, we find that the \textit{Road Surface Index} (RSI) has a good correlation with the increase in stay-duration. Therefore, we conclude that the road surface index, $R_I$ could be a vital feature to characterize the different type of stay-locations. Existing literature~\cite{38nericell} have shown that data from IMU sensors like accelerometer can be used to compute RSI of a road. In particular, the z-axis of accelerometer is seen to vary within a specific range for poor road conditions. Thus, to compute the RSI, we first process the accelerometer data, obtained from the smartphone application, to orient it according to the actual z-axis and then subsequently use the z-axis value to compute the RSI adapted as follows.  

We first take a window of $50$ meters before a stay-location and capture the z-axis acceleration for that entire window, along with the velocity of the bus. Subsequently, the RSI can be calculated as the ratio of \emph{root mean square} (RMS) of the z-axis acceleration values and the mean of the velocity in that window.
\begin{figure}
    \centering
    \begin{minipage}{0.50\columnwidth}
        \centering
        \includegraphics[width=\columnwidth,keepaspectratio]{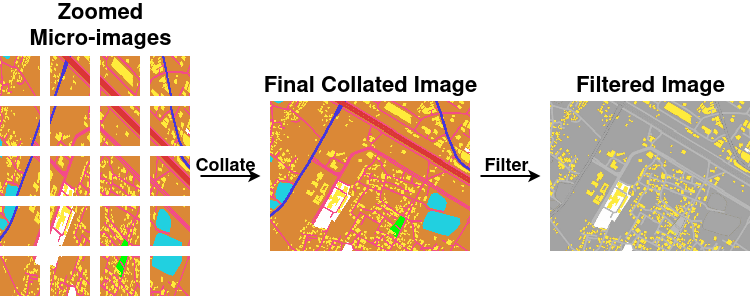}
        \vspace{2em}
        \caption{Encoding Spatial Distribution}
        \label{fig:map_process}
    \end{minipage}\hfill%
    \begin{minipage}{0.50\columnwidth}
        \centering
        \includegraphics[width=\columnwidth,keepaspectratio]{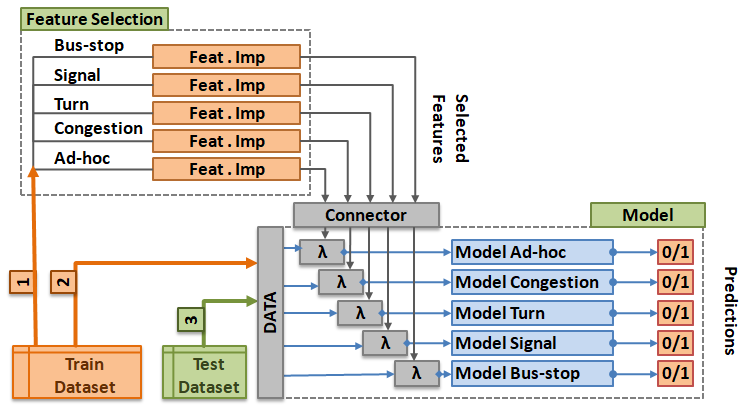}
        \caption{Architecture of the Prediction Model}
        \label{fig:model_desc}
    \end{minipage}\hfill
\end{figure}
\subsubsection{Spatial Characteristics Encoding ($\mathbf{f_{10}}$--$\mathbf{f_{13}}$)}
One of the critical deciding factors behind the choice of stay-location for a bus is its proximity to one or more landmarks. Interestingly, not only during the runtime but also in the planning stage of the bus route and its corresponding mandatory stops, the distribution of the landmarks play a vital role. This paper captures the spatial distribution of landmarks using information available from static images of the map for the area surrounding the stay-location. For this, we first capture the snapshot an area of $m \times n$ sq.m.\footnote{In this work we take $m$=$n$=$300$.} surrounding the stay-location using the Google Map API. Notably, the images directly taken from the map contain limited information regarding the area's spatial clutter. Therefore, we consider taking multiple strides over the map with a higher zoom level~\footnote{Here we fix the zoom level at $18$.}, so that they contain a detailed set of information regarding different objects (human-made or natural) present in the area. Once these individual strides are rendered completely, we subsequently collate them to create an overall detailed map of the area with more granular information. Also, during the process, we poll the API with a detailed color coding scheme so that all the different landmarks appear in different colors. Finally, when the complete detailed image is obtained, we apply filters to compute the percentage of different landmarks. A birds-eye view of this overall step is shown in Fig.~\ref{fig:map_process}.
\begin{table}[]
\centering
\scriptsize
\caption{Features Capturing Spatial Distribution of Landmarks Surrounding the Stay-location.}
\label{table:map_process}
\begin{tabular}{|l|l|}
\hline
\multicolumn{1}{|c|}{\textbf{Feature Name}} & \multicolumn{1}{c|}{\textbf{Description}}                                                                                    \\ \hline
Residential Area                            & \begin{tabular}[c]{@{}l@{}}Percentage of residential places in that area surrounding the \\ stay-location\end{tabular}       \\ \hline
Natural Land                              & \begin{tabular}[c]{@{}l@{}}Percentage of unoccupied natural land in that area \\ surrounding the stay-location.\end{tabular} \\ \hline
Road Exists &
  \begin{tabular}[c]{@{}l@{}}Percentage of road network places connecting that area of the\\ stay-location. This includes all the different types of roads like \\ one-way lanes, two-ways and highways\end{tabular} \\ \hline
Highly Populated &
  \begin{tabular}[c]{@{}l@{}}This is a binary feature denoting whether the area surrounding\\ the stay-location contains any special landmarks that may impact\\ the in and outflow of people in that area. The feature is set to $1$\\ if the area contains landmarks like schools, hospitals, public religious\\ places, parks, shopping malls, etc.\end{tabular} \\ \hline
\end{tabular}
\end{table}

Lastly, all these individual landmarks form individual features with the percentage distribution in the overall area surrounding the stay-location. The list of features extracted after this step is summarised in Table~\ref{table:map_process}. 
\subsection{Stay-location Prediction Model}
Once the features are computed, this module predicts the type of stay-location by classifying it into $5$ distinct categories: regular bus-stop, congestion, turn, signal, and ad-hoc. However, this process of classification is not straightforward. Each of these stay-locations has individual characteristics, and thus, all the features might not be equally crucial for predicting the different stay-locations. Therefore, during the training phase, we first perform a feature selection to judiciously select the essential features that can characterize a particular type of stay-location. For selecting essential features characterizing a stay-location, we use a supervised feature selection process using an ensemble classifier (number of estimators = $250$). Subsequently, once the essential features for each stay-location are obtained, we train $5$ different models (random forest-based classifier with maximum depth = $8$ and the number of estimators = $100$), each corresponding to one of the stay-locations, in an one-vs-all setup (see Fig.~\ref{fig:model_desc}). This choice of training individual models for each stay-locations helps us predict cases where a particular stay-location occurs because of multiple stay-locations (except for the ad-hoc stops which occur exclusively in this setup). For example, say a regular bus-stop is close to a traffic signal, then this approach will allow the individual models for regular bus-stop and traffic signal to predict accordingly.

Subsequently, during runtime, whenever the bus stops at a stay-location, the features are computed from the multi-modal data, and each model is polled with its respective set of selected features. Finally, the summary of each model's prediction contributes to the prediction of the type of stay-location for that corresponding stay-location.

\section{Evaluation}
\label{eval}
We evaluate the framework's performance with the in-house collected data in two steps -- (a) first with straightforward cross-validation and (b) a more realistic evaluation with test data for an entire week. Additionally, we also develop a proof-of-concept (PoC) system on top of the developed framework that can provide real-time information on bus arrival. The details follow.
\begin{figure}
    \centering
    \begin{minipage}{0.33\columnwidth}
        \centering
        \includegraphics[width=\columnwidth,keepaspectratio]{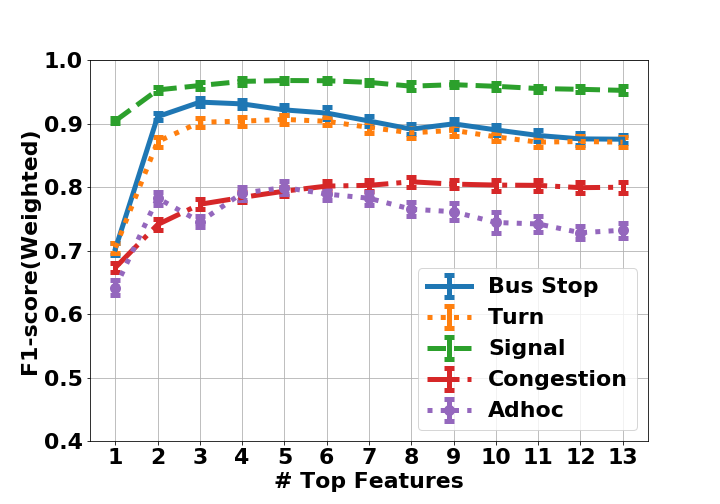}
        \subcaption{}
        \label{fig:model_perf}
    \end{minipage}\hfill
    \begin{minipage}{0.33\columnwidth}
        \centering
        \includegraphics[width=\columnwidth,keepaspectratio]{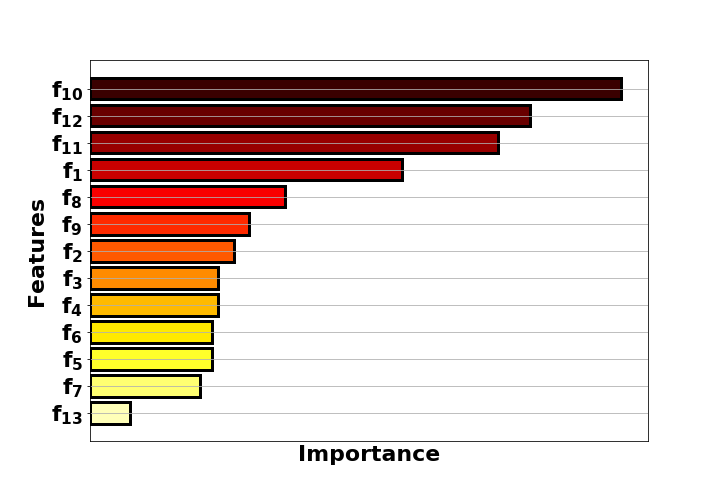}
        \subcaption{}
        \label{fig:feat_imp_bus}
    \end{minipage}\hfill
    \begin{minipage}{0.33\columnwidth}
        \centering
        \includegraphics[width=\columnwidth,keepaspectratio]{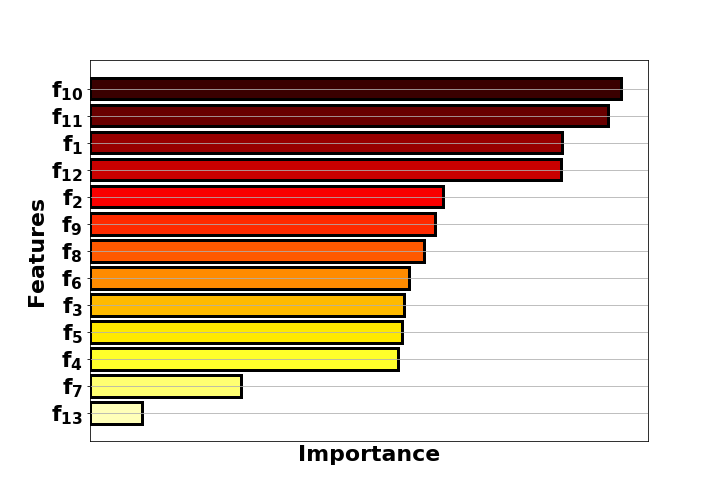}
        \subcaption{}
        \label{fig:feat_imp_adhoc}
    \end{minipage}
    \caption{(a) Performance Evaluation and Feature Importance for Individual Models -- (b) Regular Bus-stops and (c) Random Ad-hoc Stops}
\end{figure}
\subsection{Performance Evaluation}
We first evaluate the framework's performance on the collected data by performing a $5$-folds cross-validation with stratified random sampling (repeated $10$ times) in a one-vs-all setting. This approach of developing independent models for each of the stay-location types allows us to detect all the stay-location types in confounding cases. From the Fig.~\ref{fig:model_perf}, we can observe that most of these individual models achieve a maximum weighted F\textsubscript{1}-score $\ge 0.75$.

Besides, this we also perform a per model-based supervised feature selection to judiciously select the set of features that may characterize a particular stay-location more appropriately. Interestingly, from Fig.~\ref{fig:model_perf} we can see that none of the models need more than top-$8$ features to attain the maximum performance.
\begin{figure}
    \centering
    \begin{minipage}{0.33\columnwidth}
        \centering
        \includegraphics[width=\columnwidth,keepaspectratio]{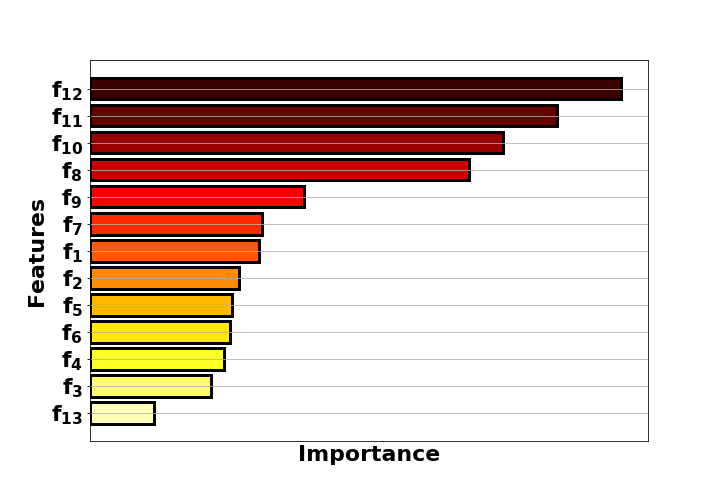}
        \subcaption{}
        \label{fig:feat_imp_signal}
    \end{minipage}\hfill
    \begin{minipage}{0.33\columnwidth}
        \centering
        \includegraphics[width=\columnwidth,keepaspectratio]{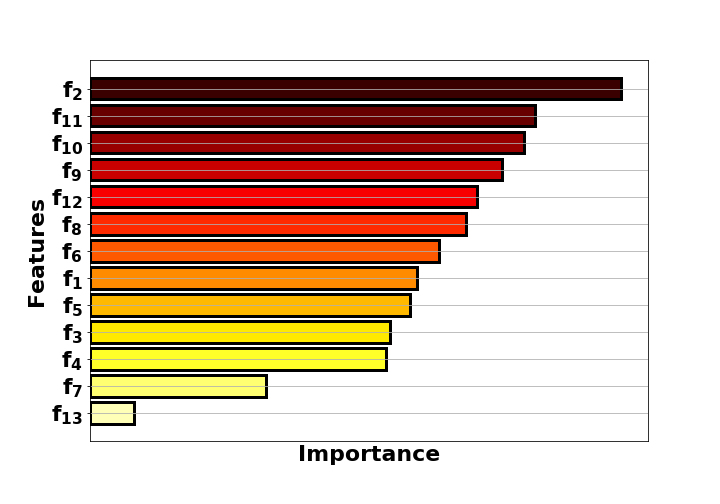}
        \subcaption{}
        \label{fig:feat_imp_congestion}
    \end{minipage}\hfill
    \begin{minipage}{0.33\columnwidth}
        \centering
        \includegraphics[width=\columnwidth,keepaspectratio]{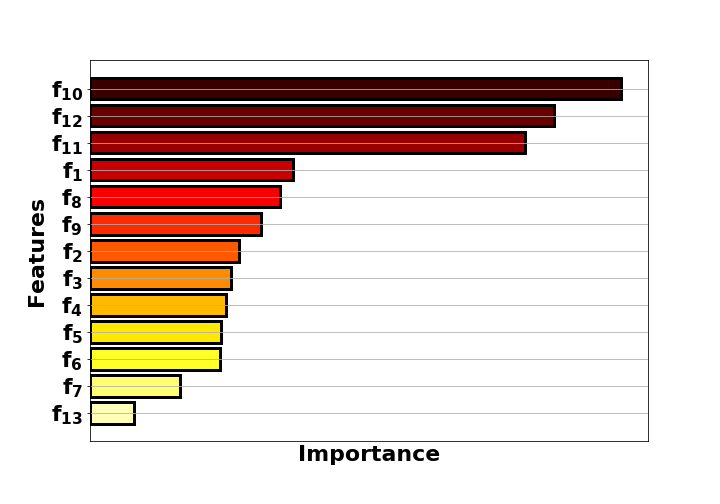}
        \subcaption{}
        \label{fig:feat_imp_turn}
    \end{minipage}
    \caption{Feature Importance for Individual Models -- (a) Traffic Signal (b) Traffic Congestion, and (c) Turn}
\end{figure}
Subsequently, we investigate further to understand the features that appear as crucial for the individual models. For regular bus-stops, we observe that spatial-encoding-based features, stay-duration, edge WiFi count (see Fig.~\ref{fig:feat_imp_bus}) appear as the top-$5$ features. This supports our initial hypothesis that regular bus-stops are usually placed in populous areas with multiple human-made structures.

Similarly, for ad-hoc stops, we see that spatial-encoding-based features, stay-duration, ambient noise-based features become more prominent (see Fig.~\ref{fig:feat_imp_adhoc}). This is because ad-hoc stops are usually given in busy areas with heavy in and outflow of the crowd. Interestingly, we observe similar patterns for stops due to congestion (see Fig.~\ref{fig:feat_imp_congestion}) where the top features include ambient noise, spatial-encoding based features, RSI, edge WiFi-count, stay-duration, which are also indicators for crowded areas along with RSI as an indicator that poor road conditions may trigger congestion. 
\begin{figure}
    \centering
    \begin{minipage}{0.50\columnwidth}
        \centering
        \includegraphics[width=\columnwidth,keepaspectratio]{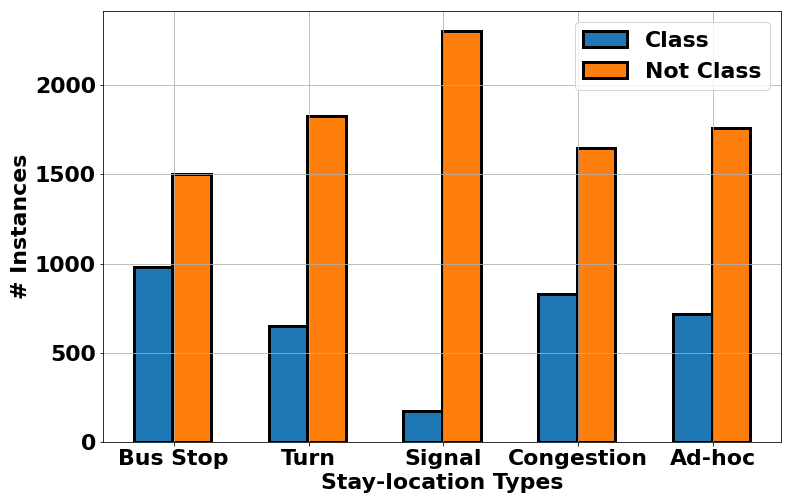}
        \subcaption{}
        \label{fig:training_distribution}
    \end{minipage}\hfill
    \begin{minipage}{0.50\columnwidth}
        \centering
        \includegraphics[width=\columnwidth,keepaspectratio]{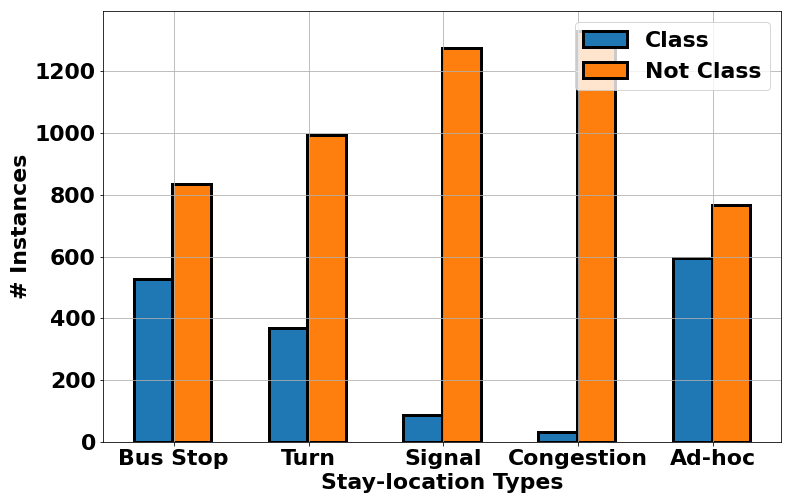}
        \subcaption{}
        \label{fig:testing_distribution}
    \end{minipage}
    \caption{Distribution of -- (a) Training Data, (b) Test Data}
\end{figure}
\begin{table}[]
\scriptsize
\centering
\caption{Impact of different Feature Groups on a Separate Test Data Prediction}
\label{tab:test_perf}
\begin{tabular}{|c|c|c|c|}
\hline
\textbf{Stay-location Type} & \textbf{Spatial Features only} & \textbf{Temporal Features only} & \textbf{\emph{BuStop}} \\ \hline
\textbf{Bus-stop}   & $0.82$      & $0.67$          & $\mathbf{0.84}$ \\ \hline
\textbf{Turn}       & $0.82$      & $0.61$          & $\mathbf{0.83}$ \\ \hline
\textbf{Signal}     & $0.82$      & $0.79$          & $\mathbf{0.92}$ \\ \hline
\textbf{Congestion} & $0.71$ & $\mathbf{0.87}$ & $0.86$          \\ \hline
\textbf{Ad-hoc}     & $0.70$ & $0.60$          & $\mathbf{0.71}$ \\ \hline
\end{tabular}
\end{table}
Finally we observe that for other stay locations like signal and turns (see Fig.~\ref{fig:feat_imp_signal} and Fig.~\ref{fig:feat_imp_signal}) features like spatial-encoding based features, edge WiFi-count, RSI and stay-duration appear as important features. It is important to note that the edge WiFi count has been more prominent in several of these cases than the WiFi count observed at the stay-location. This implies that the locality's population density through which the bus is traversing can be a good indicator of the type of stay-location that may be observed next.

Based on the understanding that we gain from this initial analysis of the framework, we subsequently analyse the performance of the framework on a test dataset collected over an entire week. Subsequently, we use this dataset to evaluate a proof-of-concept setup designed to generate the expected arrival time at any given regular bus-stop based on the output of the framework. The details follow.
\subsection{Evaluation on Test Dataset}
We first start by analyzing the data distribution across the training and the test dataset. From the Fig.~\ref{fig:training_distribution}, we observe that there is a significant imbalance in the class labels in the one-vs-all setting. Therefore, to evaluate the framework in a principled way, we first balance the training dataset for all the models using \emph{Synthetic Minority Oversampling Method} (SMOTE)~\cite{smote} and then use that dataset, after selecting the essential features, to train the corresponding models. Furthermore, we observe a significant imbalance in the testing data as well; therefore, we choose to evaluate all the models using the standard metric of weighted F~\textsubscript{1} score.

It is comforting for us to know that the framework achieves a significant accuracy in detecting the individual stay-locations with the models for regular bus-stop, ad-hoc stops, stops due to congestion, stops due to turns, and stops at traffic signal achieving a test accuracy (measured as weighted F~\textsubscript{1}) of $0.84$, $0.71$, $0.86$, $0.83$ and $0.92$ respectively. In this context, it can be observed the model for detecting ad-hoc stops performs poorer in comparison to the other models, however, the primary reason behind this is the overall random nature of these stops which is mostly guided by the availability of passengers at any given location. 

Additionally, we compare the impact of various feature groups like spatial features (all encoded features from map and RSI) with temporally varying features (WiFi count, ambient noise, stay-duration). From Table~\ref{tab:test_perf}, we observe that spatially dependent stay-locations like bus-stops, signals, and turns can be predicted with appreciable accuracy even with spatial features only; however, the prediction accuracy of temporally dependent stay-locations like traffic congestion drops significantly. Notably, our framework, which uses all these features judiciously with per-model feature selection, performs better in almost all the scenarios.
\begin{figure}
    \centering
    \begin{minipage}{0.66\columnwidth}
        \includegraphics[width=\columnwidth,keepaspectratio]{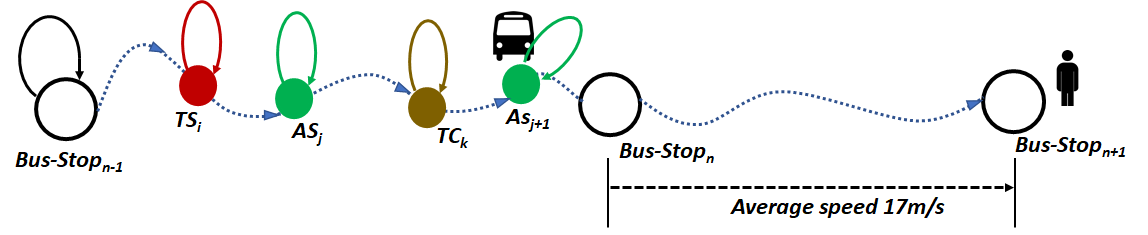}
        \vspace{1.5em}
        \caption{Realisation of the PoC Setup on Predicting Arrival Time for Public Buses}
        \label{fig:markov}
    \end{minipage}\hfill
    \begin{minipage}{0.33\columnwidth}
        \centering
        \includegraphics[width=\columnwidth,keepaspectratio]{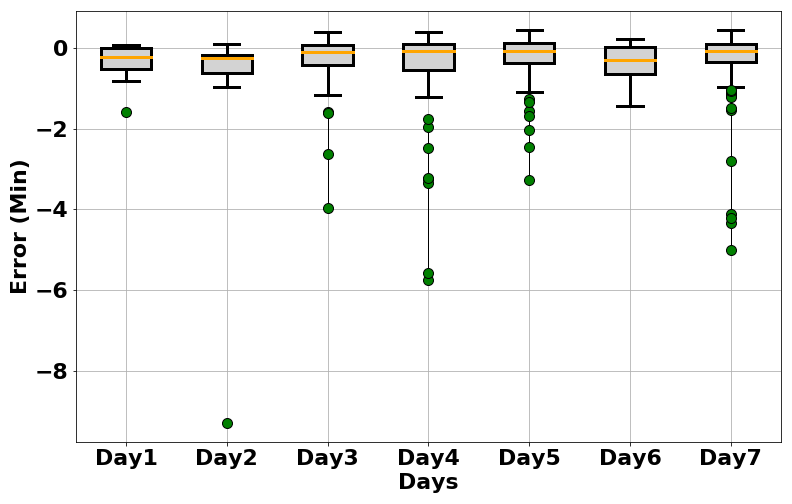}
        \caption{Day-wise Variations in Error in Predicting Expected Arrival Time}
        \label{fig:tracking_daywise}
    \end{minipage}
\end{figure}
\begin{figure}
    \centering
    \begin{minipage}{0.25\columnwidth}
        \centering
        \includegraphics[width=\columnwidth,keepaspectratio]{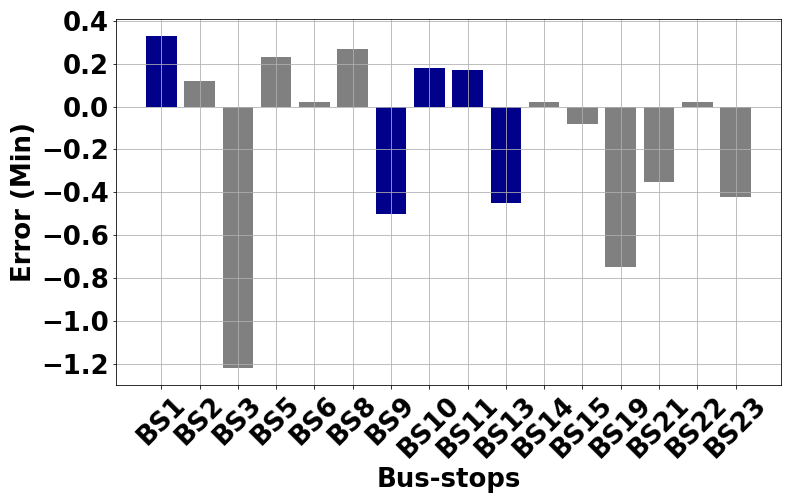}
        \subcaption{Early Morning}
    \end{minipage}\hfill
    \begin{minipage}{0.25\columnwidth}
        \centering
        \includegraphics[width=\columnwidth,keepaspectratio]{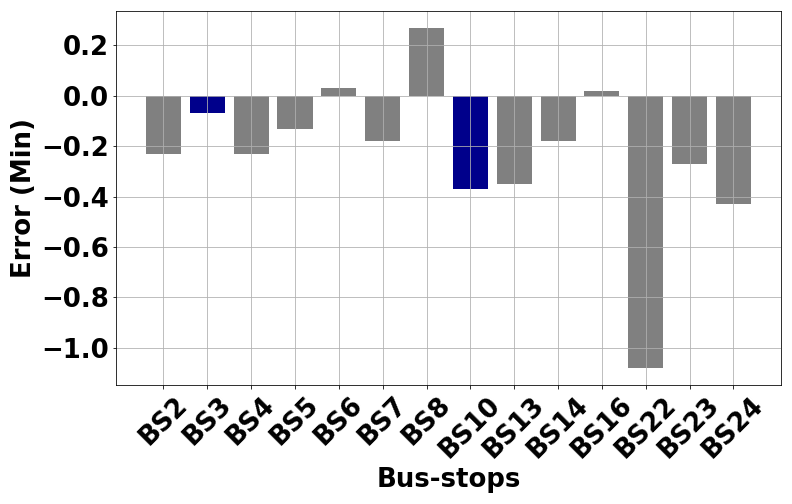}
        \subcaption{Morning}
    \end{minipage}\hfill
    \begin{minipage}{0.25\columnwidth}
        \centering
        \includegraphics[width=\columnwidth,keepaspectratio]{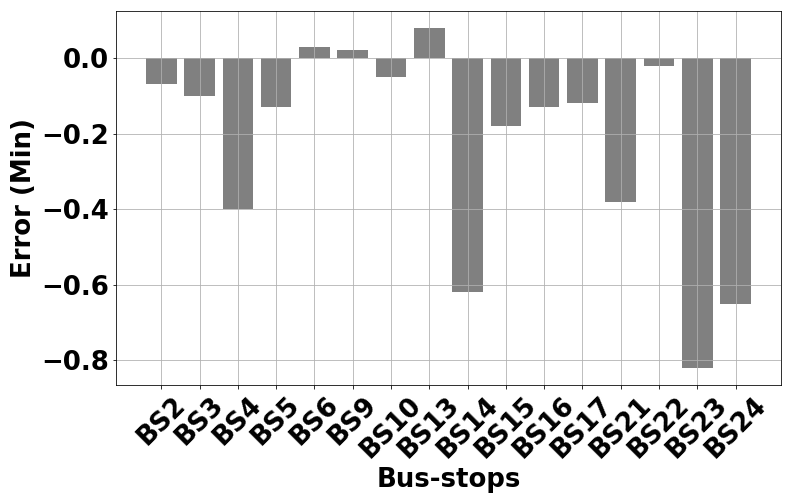}
        \subcaption{Afternoon}
    \end{minipage}\hfill
    \begin{minipage}{0.25\columnwidth}
        \centering
        \includegraphics[width=\columnwidth,keepaspectratio]{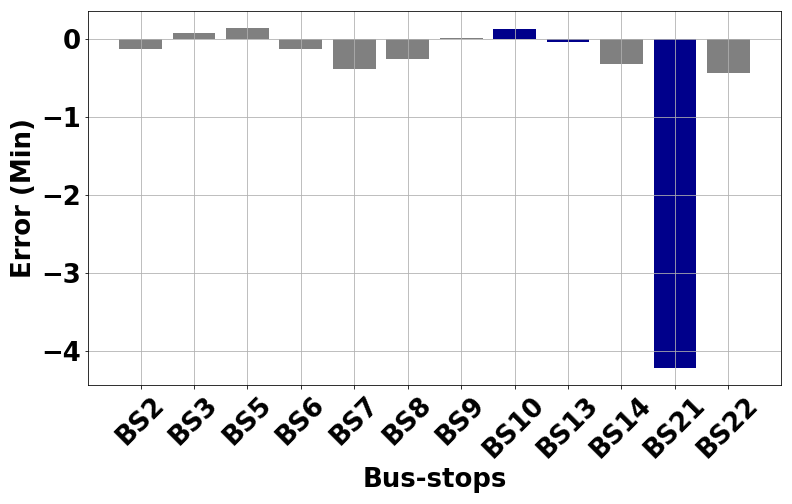}
        \subcaption{Evening}
    \end{minipage}
    \caption{Variations in Tracking Errors across Timezones -- Up Trail. The bars marked in blue are the ones predicted wrongly by the framework.}
    \label{fig:up_trail}
\end{figure}
\begin{figure}
    \centering
    \begin{minipage}{0.25\columnwidth}
        \centering
        \includegraphics[width=\columnwidth,keepaspectratio]{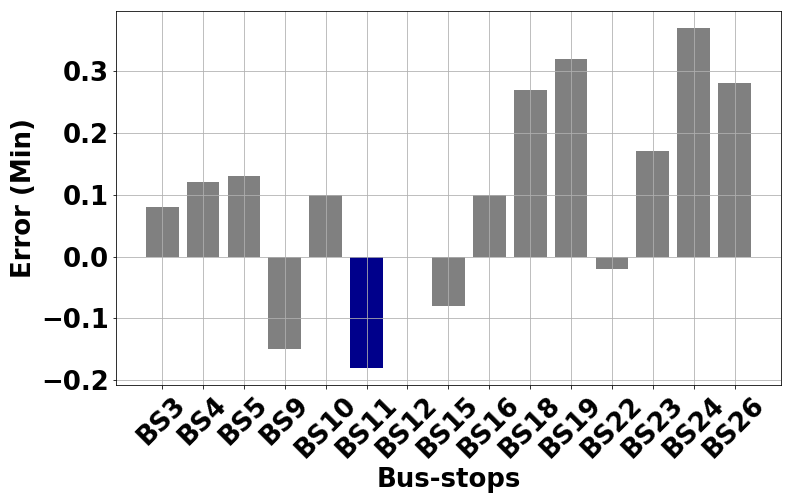}
        \subcaption{Early Morning}
    \end{minipage}\hfill
    \begin{minipage}{0.25\columnwidth}
        \centering
        \includegraphics[width=\columnwidth,keepaspectratio]{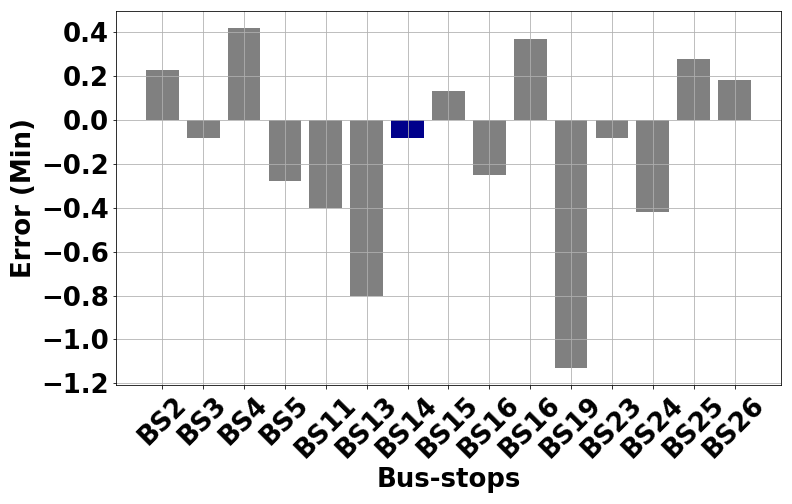}
        \subcaption{Morning}
    \end{minipage}\hfill
    \begin{minipage}{0.25\columnwidth}
        \centering
        \includegraphics[width=\columnwidth,keepaspectratio]{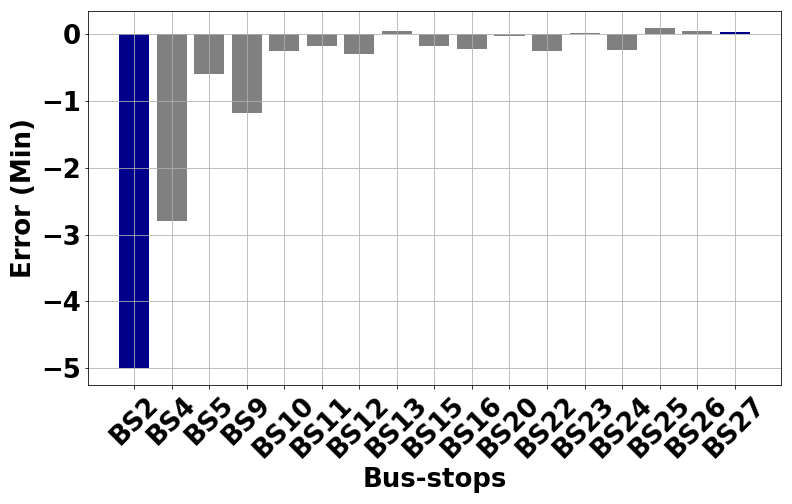}
        \subcaption{Afternoon}
    \end{minipage}\hfill
    \begin{minipage}{0.25\columnwidth}
        \centering
        \includegraphics[width=\columnwidth,keepaspectratio]{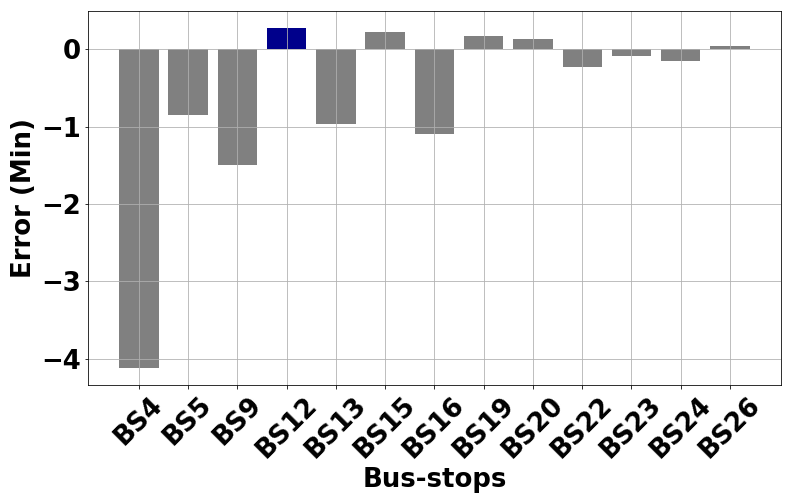}
        \subcaption{Evening}
    \end{minipage}
    \caption{Variations in Tracking Errors across Timezones. The bars marked in blue are the ones predicted wrongly by the framework. -- Down Trail}
    \label{fig:down_trail}
\end{figure}
\begin{table}[]
\scriptsize
\centering
\caption{Error (in Minutes) in Obtaining Expected Arrival Time at a Regular Bus-stops from Previously Visited Regular Bus-stops}
\label{tab:time_pred_table}
\begin{tabular}{|l|
>{\columncolor[HTML]{9B9B9B}}l |l|l|l|l|l|l|l|l|l|l|l|l|l|l|}
\hline
 &
  \cellcolor[HTML]{00D2CB}\textbf{BS1} &
  \textbf{BS2} &
  \textbf{BS3} &
  \textbf{BS4} &
  \textbf{BS5} &
  \textbf{BS6} &
  \textbf{BS7} &
  \textbf{BS8} &
  \textbf{BS9} &
  \cellcolor[HTML]{00D2CB}\textbf{BS12} &
  \textbf{BS13} &
  \textbf{BS17} &
  \cellcolor[HTML]{00D2CB}\textbf{BS18} &
  \textbf{BS19} &
  \cellcolor[HTML]{00D2CB}\textbf{BS22} \\ \hline
\cellcolor[HTML]{00D2CB}\textbf{BS1} &
   &
  -0.05 &
  -3.58 &
  -4.43 &
  -4.45 &
  -4.5 &
  -4.63 &
  -4.8 &
  -5.38 &
  -7.85 &
  -8.77 &
  -10.8 &
  -14.87 &
  -16.93 &
  -18.75 \\ \hline
\textbf{BS2} &
   &
  \cellcolor[HTML]{9B9B9B} &
  -3.53 &
  -4.38 &
  -4.4 &
  -4.45 &
  -4.58 &
  -4.75 &
  -5.33 &
  -7.8 &
  -8.72 &
  -10.75 &
  -14.82 &
  -16.88 &
  -18.7 \\ \hline
\textbf{BS3} &
   &
  \cellcolor[HTML]{9B9B9B} &
  \cellcolor[HTML]{9B9B9B} &
  -0.85 &
  -0.87 &
  -0.92 &
  -1.05 &
  -1.22 &
  -1.8 &
  -4.27 &
  -5.18 &
  -7.22 &
  -11.28 &
  -13.35 &
  -15.17 \\ \hline
\textbf{BS4} &
   &
  \cellcolor[HTML]{9B9B9B} &
  \cellcolor[HTML]{9B9B9B} &
  \cellcolor[HTML]{9B9B9B} &
  -0.02 &
  -0.07 &
  -0.2 &
  -0.37 &
  -0.95 &
  -3.42 &
  -4.33 &
  -6.37 &
  -10.43 &
  -12.5 &
  -14.32 \\ \hline
\textbf{BS5} &
   &
  \cellcolor[HTML]{9B9B9B} &
  \cellcolor[HTML]{9B9B9B} &
  \cellcolor[HTML]{9B9B9B} &
  \cellcolor[HTML]{9B9B9B} &
  -0.05 &
  -0.18 &
  -0.35 &
  -0.93 &
  -3.4 &
  -4.32 &
  -6.35 &
  -10.42 &
  -12.48 &
  -14.3 \\ \hline
\textbf{BS6} &
   &
  \cellcolor[HTML]{9B9B9B} &
  \cellcolor[HTML]{9B9B9B} &
  \cellcolor[HTML]{9B9B9B} &
  \cellcolor[HTML]{9B9B9B} &
  \cellcolor[HTML]{9B9B9B} &
  -0.13 &
  -0.3 &
  -0.88 &
  -3.35 &
  -4.27 &
  -6.3 &
  -10.37 &
  -12.43 &
  -14.25 \\ \hline
\textbf{BS7} &
   &
  \cellcolor[HTML]{9B9B9B} &
  \cellcolor[HTML]{9B9B9B} &
  \cellcolor[HTML]{9B9B9B} &
  \cellcolor[HTML]{9B9B9B} &
  \cellcolor[HTML]{9B9B9B} &
  \cellcolor[HTML]{9B9B9B} &
  -0.17 &
  -0.75 &
  -3.22 &
  -4.13 &
  -6.17 &
  -10.23 &
  -12.3 &
  -14.12 \\ \hline
\textbf{BS8} &
   &
  \cellcolor[HTML]{9B9B9B} &
  \cellcolor[HTML]{9B9B9B} &
  \cellcolor[HTML]{9B9B9B} &
  \cellcolor[HTML]{9B9B9B} &
  \cellcolor[HTML]{9B9B9B} &
  \cellcolor[HTML]{9B9B9B} &
  \cellcolor[HTML]{9B9B9B} &
  -0.58 &
  -3.05 &
  -3.97 &
  -6 &
  -10.07 &
  -12.13 &
  -13.95 \\ \hline
\textbf{BS9} &
   &
  \cellcolor[HTML]{9B9B9B} &
  \cellcolor[HTML]{9B9B9B} &
  \cellcolor[HTML]{9B9B9B} &
  \cellcolor[HTML]{9B9B9B} &
  \cellcolor[HTML]{9B9B9B} &
  \cellcolor[HTML]{9B9B9B} &
  \cellcolor[HTML]{9B9B9B} &
  \cellcolor[HTML]{9B9B9B} &
  -2.47 &
  -3.38 &
  -5.42 &
  -9.48 &
  -11.55 &
  -13.37 \\ \hline
\cellcolor[HTML]{00D2CB}\textbf{BS12} &
   &
  \cellcolor[HTML]{9B9B9B} &
  \cellcolor[HTML]{9B9B9B} &
  \cellcolor[HTML]{9B9B9B} &
  \cellcolor[HTML]{9B9B9B} &
  \cellcolor[HTML]{9B9B9B} &
  \cellcolor[HTML]{9B9B9B} &
  \cellcolor[HTML]{9B9B9B} &
  \cellcolor[HTML]{9B9B9B} &
  \cellcolor[HTML]{9B9B9B} &
  -0.92 &
  -2.95 &
  -7.02 &
  -9.08 &
  -10.9 \\ \hline
\textbf{BS13} &
   &
  \cellcolor[HTML]{9B9B9B} &
  \cellcolor[HTML]{9B9B9B} &
  \cellcolor[HTML]{9B9B9B} &
  \cellcolor[HTML]{9B9B9B} &
  \cellcolor[HTML]{9B9B9B} &
  \cellcolor[HTML]{9B9B9B} &
  \cellcolor[HTML]{9B9B9B} &
  \cellcolor[HTML]{9B9B9B} &
  \cellcolor[HTML]{9B9B9B} &
  \cellcolor[HTML]{9B9B9B} &
  -2.03 &
  -6.1 &
  -8.17 &
  -9.98 \\ \hline
\textbf{BS17} &
   &
  \cellcolor[HTML]{9B9B9B} &
  \cellcolor[HTML]{9B9B9B} &
  \cellcolor[HTML]{9B9B9B} &
  \cellcolor[HTML]{9B9B9B} &
  \cellcolor[HTML]{9B9B9B} &
  \cellcolor[HTML]{9B9B9B} &
  \cellcolor[HTML]{9B9B9B} &
  \cellcolor[HTML]{9B9B9B} &
  \cellcolor[HTML]{9B9B9B} &
  \cellcolor[HTML]{9B9B9B} &
  \cellcolor[HTML]{9B9B9B} &
  -4.07 &
  -6.13 &
  -7.95 \\ \hline
\cellcolor[HTML]{00D2CB}\textbf{BS18} &
   &
  \cellcolor[HTML]{9B9B9B} &
  \cellcolor[HTML]{9B9B9B} &
  \cellcolor[HTML]{9B9B9B} &
  \cellcolor[HTML]{9B9B9B} &
  \cellcolor[HTML]{9B9B9B} &
  \cellcolor[HTML]{9B9B9B} &
  \cellcolor[HTML]{9B9B9B} &
  \cellcolor[HTML]{9B9B9B} &
  \cellcolor[HTML]{9B9B9B} &
  \cellcolor[HTML]{9B9B9B} &
  \cellcolor[HTML]{9B9B9B} &
  \cellcolor[HTML]{9B9B9B} &
  -2.07 &
  -3.88 \\ \hline
\textbf{BS19} &
   &
  \cellcolor[HTML]{9B9B9B} &
  \cellcolor[HTML]{9B9B9B} &
  \cellcolor[HTML]{9B9B9B} &
  \cellcolor[HTML]{9B9B9B} &
  \cellcolor[HTML]{9B9B9B} &
  \cellcolor[HTML]{9B9B9B} &
  \cellcolor[HTML]{9B9B9B} &
  \cellcolor[HTML]{9B9B9B} &
  \cellcolor[HTML]{9B9B9B} &
  \cellcolor[HTML]{9B9B9B} &
  \cellcolor[HTML]{9B9B9B} &
  \cellcolor[HTML]{9B9B9B} &
  \cellcolor[HTML]{9B9B9B} &
  -1.82 \\ \hline
\cellcolor[HTML]{00D2CB}\textbf{BS22} &
   &
  \cellcolor[HTML]{9B9B9B} &
  \cellcolor[HTML]{9B9B9B} &
  \cellcolor[HTML]{9B9B9B} &
  \cellcolor[HTML]{9B9B9B} &
  \cellcolor[HTML]{9B9B9B} &
  \cellcolor[HTML]{9B9B9B} &
  \cellcolor[HTML]{9B9B9B} &
  \cellcolor[HTML]{9B9B9B} &
  \cellcolor[HTML]{9B9B9B} &
  \cellcolor[HTML]{9B9B9B} &
  \cellcolor[HTML]{9B9B9B} &
  \cellcolor[HTML]{9B9B9B} &
  \cellcolor[HTML]{9B9B9B} &
  \cellcolor[HTML]{9B9B9B} \\ \hline
\end{tabular}
\end{table}
\subsection{Proof-of-Concept Real-time Tracking of Public-bus}
One of the primary motivations behind developing~\emph{BuStop} is providing an overall service for real-time tracking of public buses. As explained in previous sections, this real-time characterizing of stay-locations fine-tunes the tracking and provides the commuters with updated information about the bus's arrival at their intended regular bus-stops. Therefore, based on this idea, we develop a simple PoC system on top of~\emph{BuStop} and simulate the system over the test dataset. The details of the system and, subsequently, its evaluation follow.

We firstly realize the entire system as a simple Markov model (as shown in Fig.~\ref{fig:markov}) whereby the arrival at any $l$\textsuperscript{th} stay-location depends on the arrival at its previous $(l-1)$\textsuperscript{th} stay-location, the stay-duration at the previous stay-location depending on its characteristics and the travel time between the two stay-locations. For example, for the location Bus-Stop\textsubscript{$n$} the arrival time depends on the nature of the previous ad-hoc stop AS\textsubscript{$j+1$}, and for the Bus-Stop\textsubscript{$n+1$} it depends on Bus-Stop\textsubscript{$n$}. Notably, any stay-location can be any one of the $5$ different types that our framework detects. However, as the stay-duration in these stay-locations may vary depending on the timezone (as shown in Section~\ref{pilot}), therefore, we obtain the mean stay-duration for each timezone from the previously seen training data and use it for the simulation. Additionally, we assume that the bus travels at an average speed of $17$m/s, a value obtained empirically after analyzing the training data.

The analysis of this setup on the test dataset reveals that the overall deviation in predicting the arrival-time at a regular bus-stop by analyzing its previous stay-location does not vary much across different days of the week (shown in Fig.~\ref{fig:tracking_daywise}). It is further comforting for us to know that the median deviation for almost all the cases is very appreciably close to $0$, which establishes the claim of accurately predicting the arrival time. 

Next, we investigate this behavior and perform a granular analysis on the separate 18 up trails (from the bus terminus to railway station) and 19 down trails (from railway station to bus terminus). From this detailed analysis (shown in Fig.~\ref{fig:up_trail} and Fig.~\ref{fig:down_trail}), we observe that for both up and down trails, the overall deviation is more during early morning and afternoon in comparison to morning and evening. This is because during early morning and afternoon the number of commuters decreases significantly due to which the buses usually move slowly intending to get more passengers and thus deviating a lot from the average speed that we have considered for the prediction. Notably, we also observe that in none of the trails, the bus stops at all the regular bus-stops~\footnote{The actual number of regular bus-stops in up and down the trail are 25 and 28 respectively.} which also impacts the performance of this system to some extent. However, we consider this a problem of the prevailing law and order of the city transit, which is beyond the scope of this paper's discussion.

Although the aforementioned setup analyses this PoC system's accuracy in Markovian setup, however, in real-time, a commuter can be present in any of the regular bus-stops where the bus has not arrived yet and enquire about the expected arrival time. Understanding this, we analyze the setup by calculating the expected arrival time at any regular bus-stop from any of the previous $(n-1)$ regular bus-stops. From the result shown in Table~\ref{tab:time_pred_table}, we observe that the error is high when the expected arrival time at a regular bus-stop is obtained when the bus is far. Nevertheless, as the bus arrives closer to the intended stop, the deviation reduces. During this analysis, we also observe that in some cases, the~\emph{BuStop} framework fails to correctly predict a regular bus-stop (the cells marked in blue). In such cases, the final error because of the wrong characterization accumulates, resulting in a more significant deviation from the actual ground-truth arrival time. However, in this context, another important factor contributing to the error, in this case, is the average speed used for the overall simulation. As the speed may vary due to several factors, including driver behavior, average speed causes the expected time to be over or underestimated by the PoC setup. Notably, in real-time, this problem will not exist as the speed information will be available in that case for more accurate prediction.
\section{Discussion and Future Directions}
\label{discussion}
Understanding the different contextual information available surrounding a stay-location and the overall impact of multi-modal sensing, we present a few critical points for discussion as follows.
\subsection{Bus Arrival Time Prediction}
In this paper, we explore the possibility of developing a system on top of~\emph{BuStop} which can provide expected arrival time information to the user. It is comforting for us to know that the results reflected the framework's potential in accurately predicting the arrival time and keeping the deviation to less than $60$ seconds in most cases. Although this is a complete offline analysis simulated on the test dataset, we envision an improved performance of this setup in real-time. The reason for such an improvement is that the setup will not have to rely on average speed and obtain the current speed in real-time. This, in turn, will reduce the errors that the current setup had due to over(under) estimation of the speed. However, there are also several challenges in such a real-time deployment which include system design-related challenges like the choice of deploying~\emph{BuStop} on-device and subsequent challenges with designing the crowdsensing setup in-the-wild. All this needs further understanding and research, which we plan to complete in future versions of the work. 
\subsection{Domain Dependence}
Although the framework we develop utilizes a generic set of features that can characterize the stay-locations irrespective of the route or city. However, one important observation that we make during this phase is the general domain-dependent nature of the framework. For example, a busy bus route with intermediate points having market places, shopping malls, schools, etc., will have an entirely different set of spatial clutter than a bus route going through a less busy route. Thus, the stay-locations which have more dependency on the spatial features may be characterized in an entirely different way on these two different routes. These dependencies on the routes' (or cities') inherent characteristics may impact the model's overall performance if we try to train the model with data from a dataset and test it on a dataset from an entirely different city or route.

Although this may seem to limit the applicability of the framework, however, in practice this can be avoided. As we intend to provide digitized bus service across several routes, one can easily have a simple per-route model with a specified training phase for proper deployment. Notably, our framework is based on lightweight machine-learning models that may also ease the overall deployment cost.
\subsection{Real-time Traffic Maps}
Recent developments in online maps not only provide us with advanced route recommendations but also provide granular information regarding the various obstacles that may appear in any particular route~\cite{realmaps}. For example, a real-time traffic map can provide the traffic flow details and whether there is congestion in any given route and, if so, the granular information regarding it. On the other hand, the maps we use to determine the spatial encoding of an area surrounding a stay-location are static and do not include any information on the current traffic patterns. Although it may seem that the inclusion of real-time traffic maps would have been more realistic, the primary reason we chose static maps is that we want to understand the spatial clutter surrounding the stay-locations rather than the traffic flow. This allows us to make our system more generic, even concerning occasional traffic fluctuations, yet still effectively detecting temporally variant stay-locations like congestion.
\section{Conclusion}
Most of the existing techniques providing real-time information on public city-transportation systems depend primarily on the unimodal GPS or cellular-based localization and mobility tracking techniques that fail to provide accurate information required by the commuter pre-plan her travel. This is because these unimodal schemes only capture coarse-grain information about mobility and are limited in characterizing the stay-locations. This, in turn, affects the system's overall accuracy while predicting the expected arrival time of the user as the system lacks information on why the vehicle has stopped at some stay-location. In this paper, we propose a machine-learning driven context-aware framework named~\emph{BuStop} which can detect different types of stay-locations of a public bus, namely -- a regular bus stop, stop at a traffic signal, a stop due to excessive traffic congestion, stops due to turns on the road and finally the randomly given ad-hoc stops. The framework does this by correctly identifying and choosing context-aware features extracted from multiple modalities that allow the framework to discern between these different types of stay-locations. Rigorous evaluation of the framework on the in-house collected dataset shows that the framework can detect different types of stay locations with appreciable accuracy, thus providing an efficient way of characterizing the stay-locations. Additionally, we also develop a PoC system on top of the developed framework to analyze and identify the framework's potential in providing an accurate expected time of arrival, one of the most critical pieces of information required for pre-planning the travel. Further analysis of the PoC setup, with simulation over the test dataset, shows that the stay-locations' characterization allows the setup to predict the arrival time with a deviation of less than $60$ seconds.

%\begin{acks}
%To Robert, for the bagels and explaining CMYK and color spaces.
%\end{acks}

\bibliographystyle{ACM-Reference-Format}
\bibliography{ref/reference}

\end{document}